\documentclass[11pt,a4paper]{article}
\usepackage[hyperref]{acl2021}
\usepackage{times}
\usepackage{latexsym}
\usepackage{subcaption}
\usepackage{graphicx}
\graphicspath{ {images/} }
\usepackage{makecell}
\usepackage{changepage}
\usepackage{hhline}
\usepackage{colortbl}
\usepackage{multirow}
\usepackage{tabularx}
\usepackage{booktabs}
\usepackage{mathtools}

\usepackage{microtype}

\aclfinalcopy 

\title{\underline{S}calable \underline{A}pproach for \underline{N}ormalizing E-commerce \underline{T}ext \underline{A}ttributes \\ (SANTA)}

\author{Ravi Shankar Mishra \\
  India Machine Learning \\
  Amazon \\
  \texttt{rmsam@amazon.com} \\\And
  Kartik Mehta \\
  India Machine Learning \\
  Amazon \\
  \texttt{kartim@amazon.com} \\\And
  Nikhil Rasiwasia \\
  India Machine Learning \\
  Amazon \\
  \texttt{rasiwasi@amazon.com} \\    
}

\date{}

\begin{document}
\maketitle
\begin{abstract}
\label{sec:abstract}
In this paper, we present \textit{SANTA}, a scalable framework to automatically normalize E-commerce attribute values (e.g. ``\textit{Win 10 Pro}") to a fixed 
set of pre-defined canonical values (e.g. ``\textit{Windows 10}"). Earlier works on attribute normalization focused on ‘fuzzy string matching’ (also referred as
syntactic matching in this paper). In this work, we first perform an extensive study of nine syntactic matching algorithms and establish that `cosine' similarity leads to best results, 
showing 2.7\% improvement over commonly used Jaccard index. Next, we argue that string similarity alone is not sufficient for attribute normalization as many surface forms require going beyond 
syntactic matching (e.g. ``\textit{720p}" and ``\textit{HD}" are synonyms). While semantic techniques like unsupervised embeddings (e.g. word2vec/fastText) have shown good results
in word similarity tasks, we observed that they perform poorly to distinguish between close canonical forms, as these close forms often occur in similar
contexts. We propose to learn token embeddings using a twin network with triplet loss. We propose an embedding learning task leveraging raw attribute values and product titles to learn 
these embeddings in a self-supervised fashion. We show that providing supervision using our proposed task improves over both syntactic and unsupervised embeddings based techniques for attribute
normalization. Experiments on a real-world attribute normalization dataset of 50 attributes show that the embeddings trained using our proposed approach obtain 2.3\% improvement over best 
string matching and 19.3\% improvement over best unsupervised embeddings.
\end{abstract}

\section{Introduction}
\label{sec:intro}

\begin{figure}
	\begin{center}
		\centerline{\includegraphics[width=0.7\columnwidth, height=6cm]{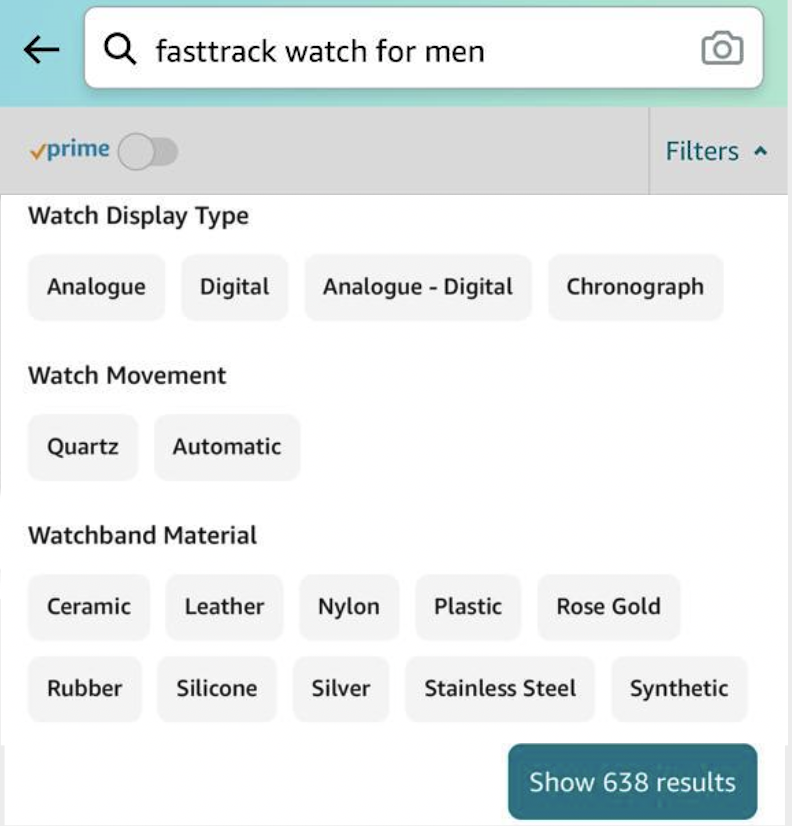}}
		\caption{Search filters widget on Amazon.}
		\label{refinement}
	\end{center}
	\vspace{-5mm}
\end{figure}

E-commerce websites like Amazon are marketplaces where multiple sellers can list and sell their products. At the time of product listing, these sellers often provide product title and structured product information
(e.g. \textit{color}), henceforth, termed as ‘product attributes’\footnote{We use the terms `product attributes' and `attributes' interchangeably in this paper.}. During the listing process, 
some attribute values have to be chosen from drop-down list (having fixed set of values to choose from) and some attributes are free-form text (allowing any value to be filled). Multiple 
sellers may express these free-form attribute values in different forms, e.g. ``\textit{HD}", ``\textit{1280 X 720}" and ``\textit{720p}" represents same TV resolution. Normalizing (or mapping) 
these raw attribute values (henceforth termed as ‘surface forms’) to same canonical form will help improve customer experience and is crucial for multiple underlying applications like search
filters, product comparison and detecting duplicates. E-commerce websites provide functionality to refine search results (refer figure~\ref{refinement}), where customers can filter based
on attribute canonical values. Choosing one of the canonical values restricts results to only those products which have the corresponding attribute value. A good normalization solution will 
ensure that products having synonym surface form (e.g. `720p' vs `HD') are not filtered out on applying such filters. 

Normalization can be considered as a two step process consisting of - a) identifying list of canonical forms for an attribute, 
and, b) mapping surface forms to one of these canonical forms. Identification task is relatively easier as most attributes have only few canonical forms 
(usually less than 10), whereas attributes can have thousands of surface forms. Hence, we focus on the mapping task in this paper, leaving identification of canonical forms as a future task to be explored.

Building an attribute normalization system for thousands\footnote{E.g. \citet{xu2019scaling} have 77K attributes only from `Sports \& Entertainment category'} of product attributes poses multiple 
challenges such as:
\begin{itemize}
	\item Presence of spelling mistakes (e.g. ``\textit{grey}" vs ``\textit{gray}", ``\textit{crom os}" vs ``\textit{chrome os}")
	\vspace{-2mm}
	\item Requirement of semantic matches (e.g. ``\textit{linux}" vs ``\textit{ubuntu}", ``\textit{mac os}" vs ``\textit{ios}")
	\vspace{-2mm}
	\item Existence of abbreviations (``\textit{polyurethane}" vs ``\textit{PU}", ``\textit{SSD}" vs ``\textit{solid state drive}")
	\vspace{-2mm}
	\item Presence of multi-token surface forms and canonical forms (e.g. ``\textit{windows 7 home}")
	\vspace{-2mm}
	\item Presence of close canonical forms (e.g. ``\textit{windows 8.1}" and ``\textit{windows 8}" can be two separate canonical forms)
\end{itemize}

Addressing these challenges in automated manner is the primary focus of this work. One can use lexical similarity of raw attribute 
value (surface form) to a list of canonical values and learn a normalization dictionary \cite{pew}. For example, lexical similarity can be used to normalize ``\textit{windows 7 home}" to ``\textit{windows 7}" or ``\textit{light blue}" to ``\textit{blue}".
However, lexical similarity-based approaches won't be able to handle cases where understanding the meaning of attribute value is important (e.g. matching ``\textit{ubuntu} to ``\textit{linux}" or ``\textit{maroon}" to ``\textit{red}").
Another alternative is to learn distributed representation (embeddings) of
surface forms and canonical forms and use similarity in embedding space for normalization. One can use unsupervised word embeddings \cite{word_similarity} 
(e.g. word2vec and fastText) for this. However, these approaches are designed to keep embeddings close by for tokens/entities
which appear in similar contexts. As we shall see, these unsupervised embeddings do a poor job at distinguishing close canonical attribute forms. 

In this paper, we describe SANTA, a scalable framework for normalizing E-commerce text attributes. Our proposed framework uses twin network \cite{siamese} with triplet loss  to learn 
embeddings of attribute values (canonical and surface forms).
We propose a self supervision task for learning these embeddings in automated manner, without requiring any manually created training data. To the best of our knowledge, 
our work is first successful attempt at creating an automated framework for E-commerce attribute normalization that can be easily extended to thousands of attributes. 

Our paper has following contributions : (1) we do a systematic study of nine lexical matching approaches for attribute normalization, (2) we propose a self supervision task for learning
embeddings of attribute surface forms and canonical forms in automated manner and describe a fully automated framework for attribute normalization using twin network and triplet loss,
and, (3) we curate an attribute normalization test set of $2500$ surface forms across $50$ attributes and present an extensive evaluation of various
approaches on this dataset. We also show an independent analysis on syntactic and semantic portions of this dataset and provide insights into benefits of our approach
over string similarity and other unsupervised embeddings. Rest of the paper is organized as follows. We do a literature survey of related fields in Section \ref{sec:related}. 
We describe string matching and embeddings based approaches, including our proposed SANTA framework, in Section~\ref{sec:overview}. We describe our experimental setup 
in Section~\ref{sec:exp} and results in Section~\ref{results}. Lastly, we summarize our work in Section~\ref{sec:conclusions}.

\section{Related Work}
\label{sec:related}
\vspace{-2mm}

\subsection{E-commerce Attribute normalization}
\vspace{-2mm}
The problem of normalizing E-commerce attribute values have received limited attention in literature. Researchers have mainly focused on normalizing 
\textit{brand} attribute, exploring combination of manual mapping curation or lexical similarity-based approaches \cite{more, pew}. \citet{more} explored use
of manually created key-value pairs for normalizing brand values extracted from product titles. \citet{pew} explored two fuzzy matching algorithms
of Jaccard similarity and Jaro-Winkler distance and found n-gram based Jaccard similarity to be performing better for brand normalization. We use this
Jaccard similarity approach as a baseline for comparison. 

\vspace{-2mm}
\subsection{Fuzzy String Matching}
\vspace{-2mm}

Fuzzy string matching has been explored for multiple applications, including address matching, names matching \cite{cohen, peter, recchia}, biomedical abbreviation matching \cite{bio} and query spelling correction. Although extensive work has been done for fuzzy string
matching, there is no consensus on which technique works best. \citet{peter} explored multiple similarity measures for personal name matching, 
and reported that best algorithm depends upon the characteristics of the dataset. \citet{cohen} experimented with edit-distance, token-based distance
and hybrid methods for matching entity names and reported best performance for a hybrid approach combining TF-IDF weighting with Jaro-Winkler distance.
\citet{recchia} did a systematic study of 21 string matching methods for the task of place name matching. While they got relatively better performance
with n-gram approaches over commonly used Levenshtein distance, they concluded that best similarity approach is task-dependent. \citet{gali} argued that
performance of the similarity measures is affected by characteristics such as text length, spelling accuracy, presence of abbreviations and underlying
language. Motivated by these learnings, we do a systematic study of fuzzy matching techniques for the problem of E-commerce attribute normalization. 
Besides, we use latest work in the field of neural embeddings for attribute normalization.

\section{Overview}
\label{sec:overview}
\begin{figure}
	\begin{center}
		\centerline{\includegraphics[width=\columnwidth, height=3.5cm]{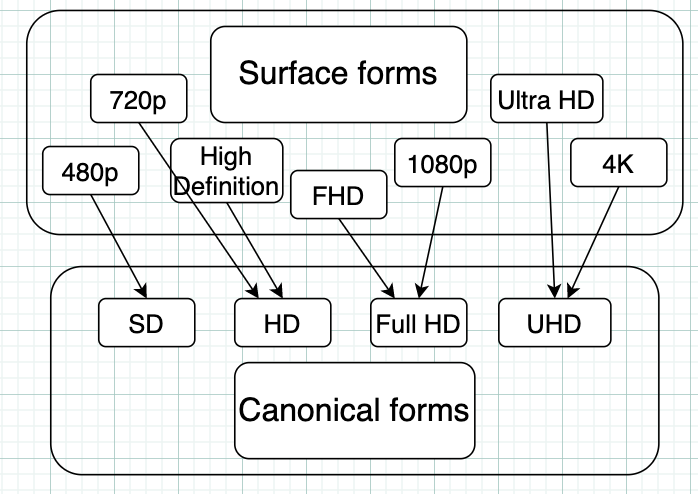}}
		\caption{Illustration of Attribute Normalization Task}
		\label{fig:Illustration}
	\end{center}
	\vspace{-3mm}
\end{figure}

Attribute normalization can be posed as a matching problem. Given an attribute surface form and a list of possible canonical forms, similarity of surface form with each 
canonical form is calculated and surface form is mapped to the canonical form with highest similarity or mapped to `other' class if none of the canonical forms is suitable 
(refer Figure~\ref{fig:Illustration} for illustration). Formally, given a surface form $s_{i}\:(i\in[1,n])$ and a list of canonical forms $c_{j}\:(j\in[0,k])$, where $c_{0}$ is the 
`other' class, $n$ is number of surface forms and $k$ is number of canonical forms. The aim is to find a mapping function $M$ such that:

\begin{equation}
\vspace{-2mm}
M(s_{i}) = c_{j}\;where \;i\in[1,n]\,, j\in[0,k]
\end{equation}

In this paper, we explore fuzzy string matching and similarity in embedding space as matching techniques. We describe multiple string matching approaches in 
Section~\ref{stringsim}, followed by unsupervised token embedding approaches in Section~\ref{unsupervisedsim} and our proposed SANTA framework in Section~\ref{supervisedsim}.

\begin{figure}
	\begin{center}
		\centerline{\includegraphics[width=\columnwidth, height=3.5cm]{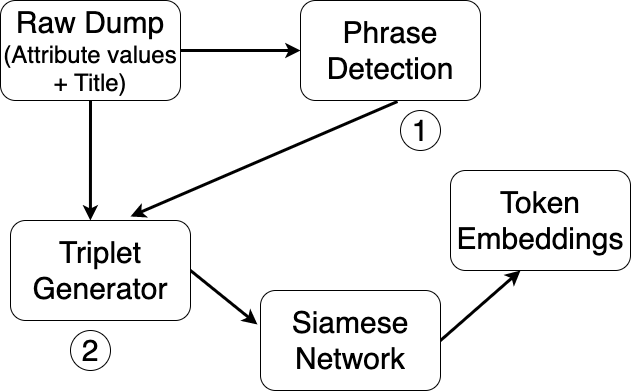}}
		\caption{SANTA framework for training embeddings suitable for attribute normalization}
		\label{fig:SANTAworkflow}
	\end{center}
	\vspace{-5mm}
\end{figure}

\vspace{-2mm}
\subsection{String Similarity Approach}
\label{stringsim}
\vspace{-2mm}

We study three different categories of string matching algorithms\footnote{https://itnext.io/string-similarity-the-basic-know-your-algorithms-guide-3de3d7346227} and explore three 
algorithms in each category\footnote{For algorithms which return distance metrics rather than similarity, we use lowest distance as substitute for highest similarity.}:

\begin{itemize}
	\item \textbf{Edit distance-based}: These algorithms compute the number of operations needed to transform one string to another, leading to higher similarity score for less operations. 
	We experimented with six algorithms in this category, a) Hamming, b) Levenshtein, and c) Jaro-Winkler.
	\vspace{-2mm}
	\item \textbf{Sequence-based}: These algorithms find common sub-sequence in two strings, leading to higher similarity score for longer common sub-sequence or a greater number of 
	common sub-sequences. We experimented with three algorithms in this category, a) longest common subsequence similarity, b) longest common substring similarity, and 
	c) Ratcliff-Obershelp similarity.
	\vspace{-2mm}
	\item \textbf{Token-based}: These algorithms represent string as set of tokens (e.g. ngrams) and compute number of common tokens between them, leading to higher similarity score for
	higher number of common tokens. We experimented with three algorithms in this category - a) Jaccard index, b) Sorensen–Dice coefficient, and c) Cosine similarity. 
	We converted strings to character ngrams of size 1 to 5 before applying this similarity.

\end{itemize}
We used python module textdistance\footnote{https://pypi.org/project/textdistance/} for all string similarity experiments. For detailed definition 
of these approaches, we refer readers to \citet{gomaa} and \citet{vijaymeena}.

\vspace{-2mm}
\subsection{Unsupervised Embeddings}
\label{unsupervisedsim}
\vspace{-2mm}
\citet{mikolov} introduced word2vec model that uses a shallow neural network to obtain distributed representation (embeddings) of words, 
ensuring words that appear in similar contexts are closer in the embedding space. To deal with unseen and rare words, \citet{mikolov1} proposed 
fastText model that improves over word2vec embeddings by considering sub-words and representing word embeddings as average of embeddings of 
corresponding sub-words.
To learn domain-specific nuances, we trained a word2vec and fastText model using a dump consisting of product titles and attribute values (refer Section~\ref{sec:exp} for details of this dump). 
We found better results with using concatenation of title with attribute value as compared to using only title, likely due to including surface form from title 
and attribute canonical form (or vice versa) in a single context.

\vspace{-2mm}
\subsection{\textbf{S}calable \textbf{A}pproach for \textbf{N}ormalizing \textbf{T}ext \textbf{A}ttributes (SANTA)}
\label{supervisedsim}
\vspace{-2mm}
Figure \ref{fig:SANTAworkflow} gives an overview of learning embeddings with our proposed SANTA framework. We define an embedding learning task using twin network 
with triplet loss to enforce that embeddings of attribute values are closer to corresponding titles as compared to embeddings of a randomly chosen title from the same 
product category. To deal with multi-word values, we use a simple step of treating each multi-word attribute value as a single phrase. Overall, we observed 40K such
phrases, e.g. ``\textit{back cover}", ``\textit{android v4.4.2}", ``\textit{9-12 month}" and ``\textit{wine red}". For both attribute values and product titles, we 
converted these multi-token phrases to single tokens (e.g. `back cover' is replaced with `back\_cover').

We describe details of the embedding learning task and triplet generation in Section~\ref{triplet_generation}, and twin network in Section~\ref{Twin_network}. 

\vspace{-2mm}
\subsubsection{Triplet Generation}
\label{triplet_generation}
\vspace{-2mm}
There are scenarios when title contains canonical form of attribute value (e.g. ``\textit{3xl}" could be size attribute value for
a title `Nike running shoes for men \underline{xxxl}'). We can leverage this information to learn embeddings that not only capture semantic similarity
but can also distinguish between close canonical forms. Motivated by work in answer selection \cite{productqna, siamese}, we define an embedding learning task of
keeping surface form closer to corresponding title as compared to a randomly chosen title. We created training data in form of triplets of
anchor ($q$), positive title ($a_{+}$) and negative title ($a_{-}$), where $q$ is attribute value, $a_{+}$ is corresponding product title and $a_{-}$ 
is a title selected randomly from product category of $a_{+}$. One way to select negatives is to pick a random product from any product category, but that may provide
limited signal for embedding learning task (e.g. choosing an Apparel category product when actual product is from Laptop category). Instead, 
we select a negative product from same product category, which acts as a hard negative \cite{kumar2019improving, facenet} and improves the attribute normalization results.
Selecting products from same category may lead to few incorrect negative titles (i.e. negative title may contain the correct attribute value). 
We screen out incorrect negatives where anchor attribute value ($q$) is mentioned in title, reducing noise in the training data.

\vspace{-2mm}
\subsubsection{Twin Network and Triplet Loss}
\label{Twin_network}
\vspace{-2mm}

\begin{figure}[ht]
	\begin{center}
		\centerline{\includegraphics[width=\columnwidth, height=6cm]{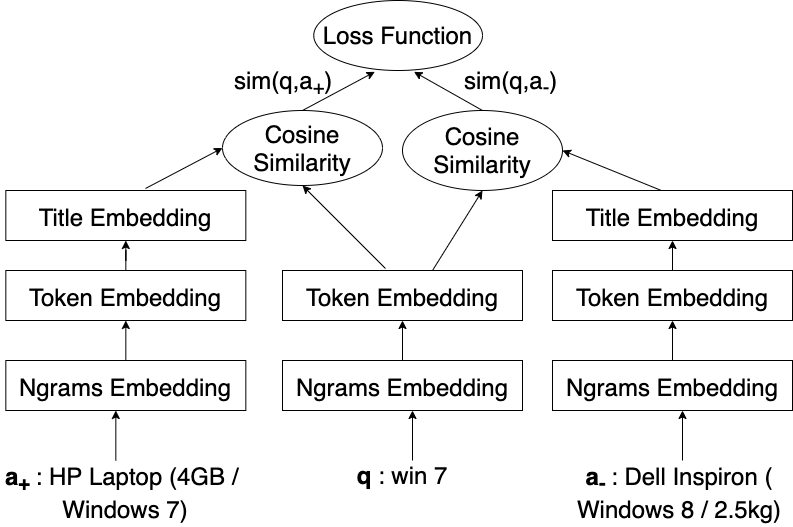}}
		\caption{Illustration of Twin network with Triplet loss.}
		\label{TripletLoss}
	\end{center}
	\vspace{-5mm}
\end{figure}

We choose twin network as it projects surface forms and canonical forms in same embedding space and triplet loss helps to keep surface forms closer 
to the most appropriate canonical form. Figure \ref{TripletLoss} describes the architecture of our SANTA framework. Given a $(q, a_{+}, a_{-})$ triplet, 
the model learns embedding that minimize the triplet loss function (Equation~\ref{eq:triplet loss}). Similar to fastText, we represent each 
token as consisting of sub-words (n-gram tokens). 
Embedding for a token is created using a composite function on sub-word embeddings, and similarly, embeddings for title are created using composite function on word embeddings. 
We use averaging of embeddings as composite function (similar to fastText), though the framework is generic and other composite functions like LSTM, CNN and transformers can also be used.

Let \emph{E} denote the embedding operator and \emph{cos} represent cosine similarity metric, then triplet loss function is given as:

\begin{equation}
\vspace{-2mm}
\begin{split}
Loss\:=\: max\:\{0, M &- cos(E(q), E(a_{+})) \\
& + cos(E(q), E(a_{-}))\}
\label{eq:triplet loss}
\end{split}
\end{equation}
where $M$ is margin.

The advantage of this formulation over unsupervised embeddings (Section~\ref{unsupervisedsim}) is that in addition to learning semantic similarities for attribute values, it also 
learns to distinguish between close canonical forms, which may appear in similar contexts. For example, the embedding of surface form `720p' will
move closer to embedding of `HD' mentioned in $a_{+}$ title but away from embedding of `Ultra HD' mentioned in $a_{-}$ title.

\section{Experimental Setup}
\label{sec:exp}

In this section, we describe our experimental setup, including dataset, metrics and hyperparameters of our model. There is no publicly available data set for 
attribute normalization problem. \citet{more} and \citet{pew} worked on brand normalization problem but the datasets are not published for reuse.
 \citet{xu2019scaling} published a dataset collected from AliExpress `Sports \& Entertainment' category for attribute extraction use-case. This dataset belongs to a single category 
 and is restricted to samples where attribute value is present in title, hence limiting its applicability for attribute normalization. 
 To ensure robust learnings, we curate a real-world attribute normalization dataset spread across multiple categories and report all our evaluations on this dataset.

\vspace{-2mm}
\subsection{Training and Test data}
\vspace{-2mm}
We selected $50$ attributes across $20$ product categories including electronics, apparel 
and furniture for our study and obtained their canonical forms from business teams. These selected attributes have on average $7.5$ canonical values (describing the exact selection
process for canonical values is outside the scope of current work). For each of these attributes, we picked top 50 surface forms 
and manually mapped these values to corresponding canonical forms, using `other'
label when none of the existing canonical forms is suitable. We, thus, obtain a labelled dataset of $2500$ samples ($50$ surface forms each for $50$ attributes),
out of which $38\%$ surface forms are mapped to `other' class. Surface forms mapping to `other' are either junk value (e.g. ``\textit{5MP}" for operating system)
or coarser value (e.g. ``\textit{android}" when canonical forms are ``\textit{android 4.1}", ``\textit{android 4.2}" etc.). It took $20$ hours of manual effort for 
creating this dataset. We split this data into two parts ($20\%$ used as dev set and $80\%$ as test set).

For training, we obtain a dump of $100K$ products corresponding to each attribute, obtaining a dump of $5M$ records ($50$ attributes X $100K$ products per attribute), 
having title and attribute values. This data ($5M$ records) is used for training unsupervised embeddings (Section~\ref{unsupervisedsim}). For each record, we 
select one negative example for triplet generation (Section~\ref{triplet_generation}) and use this triplet data (5M records) for learning SANTA model.
Kindly note that training data creation is fully automated, and does not require any manual effort, making our approach easily scalable.

\vspace{-2mm}
\subsection{Metric}
\vspace{-2mm}

There are no well-established metrics in literature for attribute normalization problem. One simple approach is to consider canonical form with highest similarity as predicted value for
evaluation. However, we argue that an algorithm should be penalized for mapping a junk value to any canonical form. Based on this motivation, we define two evaluation metrics that we use in this work.

\vspace{-2mm}
\subsubsection{Accuracy}
\vspace{-2mm}
\label{accuracy}
\begin{figure}[ht]
	\begin{center}
		\centerline{\includegraphics[width=\columnwidth, height=2cm]{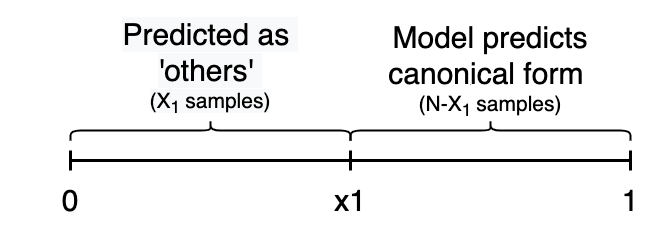}}
		\caption{Illustration for Accuracy metric.}
		\label{accuracy_metric}
	\end{center}
\end{figure}

We divide predictions on all samples ($N$) into two sets using a threshold $x_1$ (see Figure \ref{accuracy_metric}). `Other' class is predicted 
for samples having score less than $x_1$ (low similarity to any canonical form) and canonical form with highest similarity is considered for 
samples having score greater than $x_1$ (confident prediction). We consider prediction as correct for samples in $X_{1}$ set if true label is 
`other' and for samples in $N-X_1$ set, if model prediction matches the true label. We define Accuracy as ratio of correct predictions to the 
number of cases where prediction is made ($N$ in this case). The threshold $x_1$ is selected based on performance on dev set.

\vspace{-2mm}
\subsubsection{Accuracy Coverage Curve}
\vspace{-2mm}
\begin{figure}[ht]
	\begin{center}
		\centerline{\includegraphics[width=\columnwidth, height=2cm]{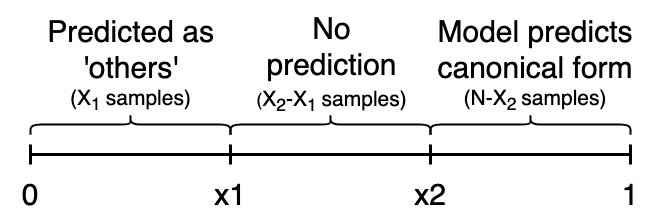}}
		\caption{Illustration for Accuracy Coverage metric.}
		\label{metric}
	\end{center}
\vspace{-5mm}
\end{figure}

It can be argued that a model is confident about surface forms when prediction score is on either extreme (close to $1$ or close to $0$). 
Motivated by this intuition, we define another metric where we divide predictions into three
sets using two thresholds $x_1$ and $x_2$ (see Figure \ref{metric}). `Other' class is predicted for samples having score less than $x_1$ (low 
similarity to any canonical form), no prediction is made for samples having score between $x_1$ and $x_2$ (model is not confidently predicting
any canonical form but confidence score is not too low to predict `other' class) and canonical form with highest similarity is considered
for samples having score greater than $x_2$. We define Coverage as fraction of samples where some prediction is made ($(X_1 + N - X_2)/N$), and 
Accuracy as ratio of correct predictions to the number of predictions. For samples in $X_{1}$ set, we consider prediction correct if true label is
`other' and for samples in $N-X_2$ set, we consider prediction correct when model prediction matches the true canonical form. The thresholds are selected based 
on performance on dev set and based on different choice of thresholds, we create Accuracy-Coverage curve for comparison.

\vspace{-2mm}
\subsection{SANTA Hyperparameters}
\vspace{-2mm}
We set the value of $M$ as $0.4$, embedding dimension as $200$, minimum n-gram size as $2$ and maximum n-gram size as $4$. We run the training
using Adadelta optimizer for 5 epochs, which took approximately $8$ hours on a NVIDIA V100 GPU. The parameters to be learned are ngram embeddings ($0.63M$ ngrams X $200$ embedding dimension 
= $127M$ parameters). Ngram embeddings are shared across the twin network.

\section{Results}
\vspace{-2mm}
\label{results}
We present systematic study on string similarity approaches in Section \ref{ss}, followed by experiments of unsupervised embeddings 
in Section \ref{FT}. We compare best results from Section \ref{ss} and Section \ref{FT} with our proposed SANTA framework in Section 
\ref{santa}. We study these algorithms separately on syntactic and semantic portion of test dataset in Section 
\ref{semanticsyntactic} and perform qualitative analysis based on t-SNE visualization in Section \ref{embedd_visual}.

\vspace{-2mm}
\subsection{Evaluation of String Similarity}
\label{ss}
\vspace{-2mm}

Table \ref{ss_approach} shows comparison of string similarity approaches for attribute normalization. We observe that token based methods performs best, 
followed by comparable performance of sequence based and edit distance based methods. We believe that 
token based approaches outperformed other approaches as 
they are insensitive to the position where common sub-string occurs in the two strings (e.g. matching ``\textit{half sleeve}" to ``\textit{sleeve half}" for \textit{sleeve 
type} attribute). \citet{pew} evaluated n-gram based `Jaccard index' (token based approach) and `Jaro-Winkler distance' (character based approach) 
for brand normalization and got similar observations, obtaining best results with `Jaccard index'. We observe that `Cosine similarity' obtains
$2.7\%$ accuracy improvement over Jaccard index in our experiments. 

\begin{table}
	\vspace{-5mm}
	\caption{Evaluation of String similarity approaches.}
	\label{ss_approach}
	\begin{center}
		\begin{small}
			\begin{sc}
				\begin{tabular}{lcc}
					\begin{tabularx}{0.8\columnwidth}{XcX}
						\toprule
						String Similarity& \makecell{Accuracy}  \\
						\midrule
						\textbf{Edit distance based} &                                                    \\
						Hamming                      & 51.6                                       \\
						Levenshtein                  & 61.1                                       \\
						Jaro-Winkler                 & 62.1                                         \\
						\midrule
						\textbf{Sequence based}      &                                                     \\
						LC Subsequence  			& 57.6                                         \\
						LC Substring    			& 64.7                                          \\
						Ratcliff-Obershelp          & 64.9                                        \\
						\midrule
						\textbf{Token based}         &                                                     \\
						Jaccard index               & 74.6                                         \\
						Sorensen-Dice   			& 74.6                                      \\
						Cosine similarity           & \textbf{76.6}                      \\
						\bottomrule
					\end{tabularx}
				\end{tabular}
			\end{sc}
		\end{small}
	\end{center}
	\vspace{-5mm}
\end{table}

\vspace{-2mm}
\subsection{Evaluation of Unsupervised Embeddings}
\label{FT}
\vspace{-2mm}
Table \ref{compare} shows performance of word2vec and fastText approach. We observe that presence of n-grams information in fastText leads to significant improvement over word2vec, 
as use of n-grams helps with matching of rare attribute values. However,  fastText is not able to match string similarity baseline (refer Table~\ref{ss_approach}).
We believe unsupervised embeddings shows relatively inferior performance for attribute normalization task, as embeddings are learnt based on contexts in product titles, 
keeping different canonical forms (e.g. ``\textit{HD}" and ``\textit{Ultra HD}") close by as they occur in similar context. 

\vspace{-2mm}
\subsection{Evaluation of SANTA framework}
\label{santa}
\vspace{-2mm}
Table \ref{compare} shows comparison of SANTA with multiple normalization approaches, including best solutions from Section~\ref{ss} and Section~\ref{FT}. 
To understand the difficulty of this task, we introduce two baselines of a) randomly mapping surface form to one of the canonical forms (termed as `RANDOM'), and b) predicting the 
most common class based on dev data (termed as `MAJORITY CLASS'). We observe 37.8\% accuracy with `RANDOM' and 48.5\% accuracy with `MAJORITY CLASS', establishing the difficulty of the task. 
SANTA (with ngrams) shows best performance with 78.4\% accuracy, leading to $2.3\%$ accuracy improvement over `Cosine Similarity' (best string similarity approach) and $19.3\%$ over 
fastText (best unsupervised embeddings). We discuss few qualitative examples for these approaches in appendix.

Figure \ref{PR_curve} shows Accuracy-Coverage curve for these algorithms. As observed from this curve, SANTA consistently outperforms string similarity and fastText across all coverages.

\begin{table}
	\vspace{2mm}
	\caption{Comparison of normalization approaches}
	\label{compare}
	\begin{center}
		\begin{small}
			\begin{sc}
				\begin{tabular}{lcc}
					\begin{tabularx}{0.85\columnwidth}{XcX}
						\toprule
						Model                        & \makecell{Accuracy} \\
						\midrule
						Random			               			& 37.8                              \\
						Majority class prediction			               			& 48.5                              \\
						\midrule
						Jaccard index               			& 74.6                                  \\
						Cosine similarity       				& 76.6                                   \\
						\midrule
						word2vec & 48.4                                     \\
						fastText               			& 65.7                                    \\
						\midrule
						SANTA (without ngrams)						& 47.4                                \\
						SANTA (with ngrams)           			& \textbf{78.4}   \\
						\bottomrule
					\end{tabularx}
				\end{tabular}
			\end{sc}
		\end{small}
	\end{center}
\end{table}

\begin{figure}[ht]
	\begin{center}
		\centerline{\includegraphics[width=\columnwidth]{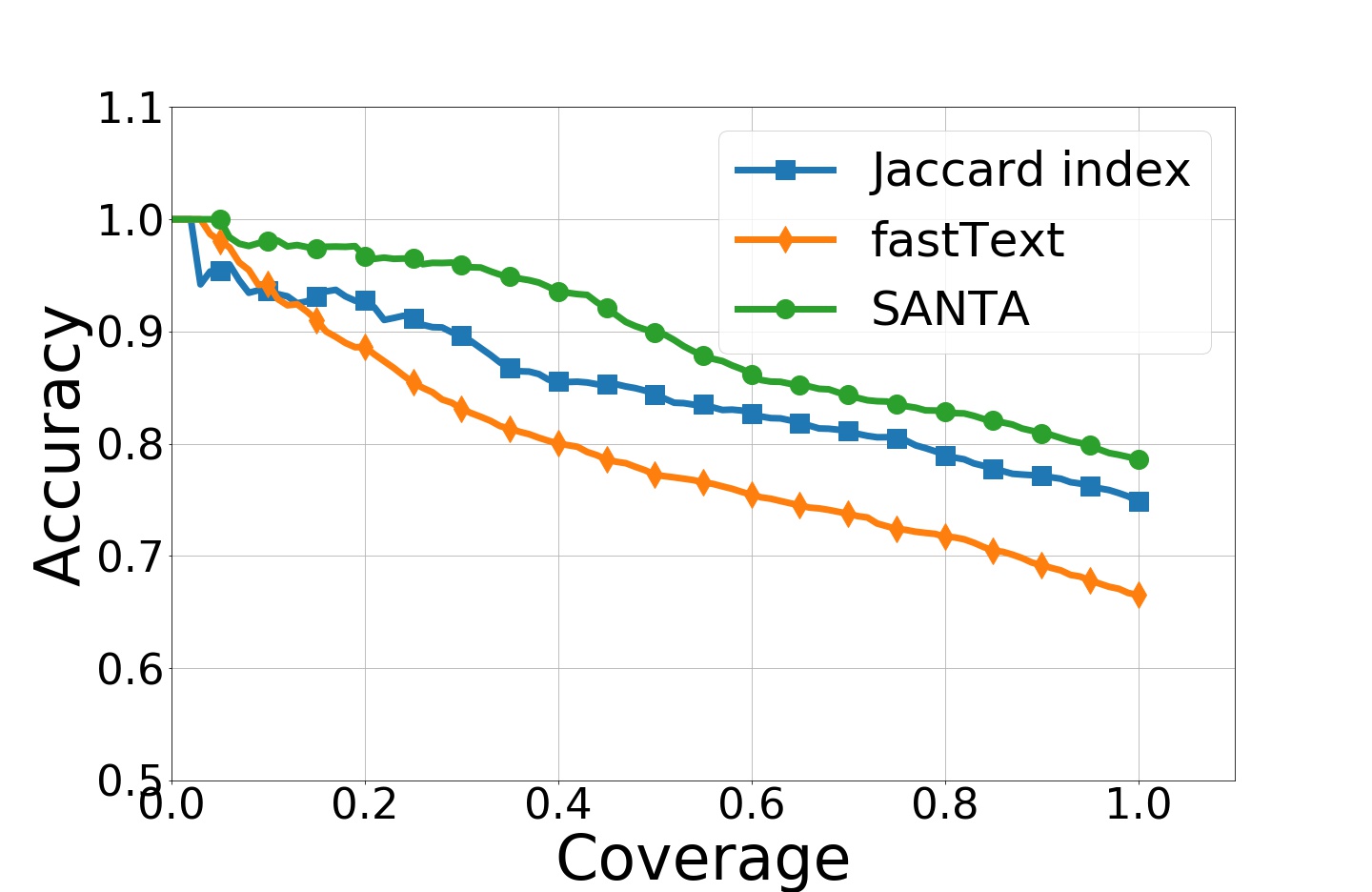}}
		\caption{Accuracy-Coverage plot for various Normalization techniques.}
		\label{PR_curve}
	\end{center}
\vspace{-7mm}
\end{figure}

\begin{figure}[ht]
	\begin{center}
		\centerline{\includegraphics[width=\columnwidth]{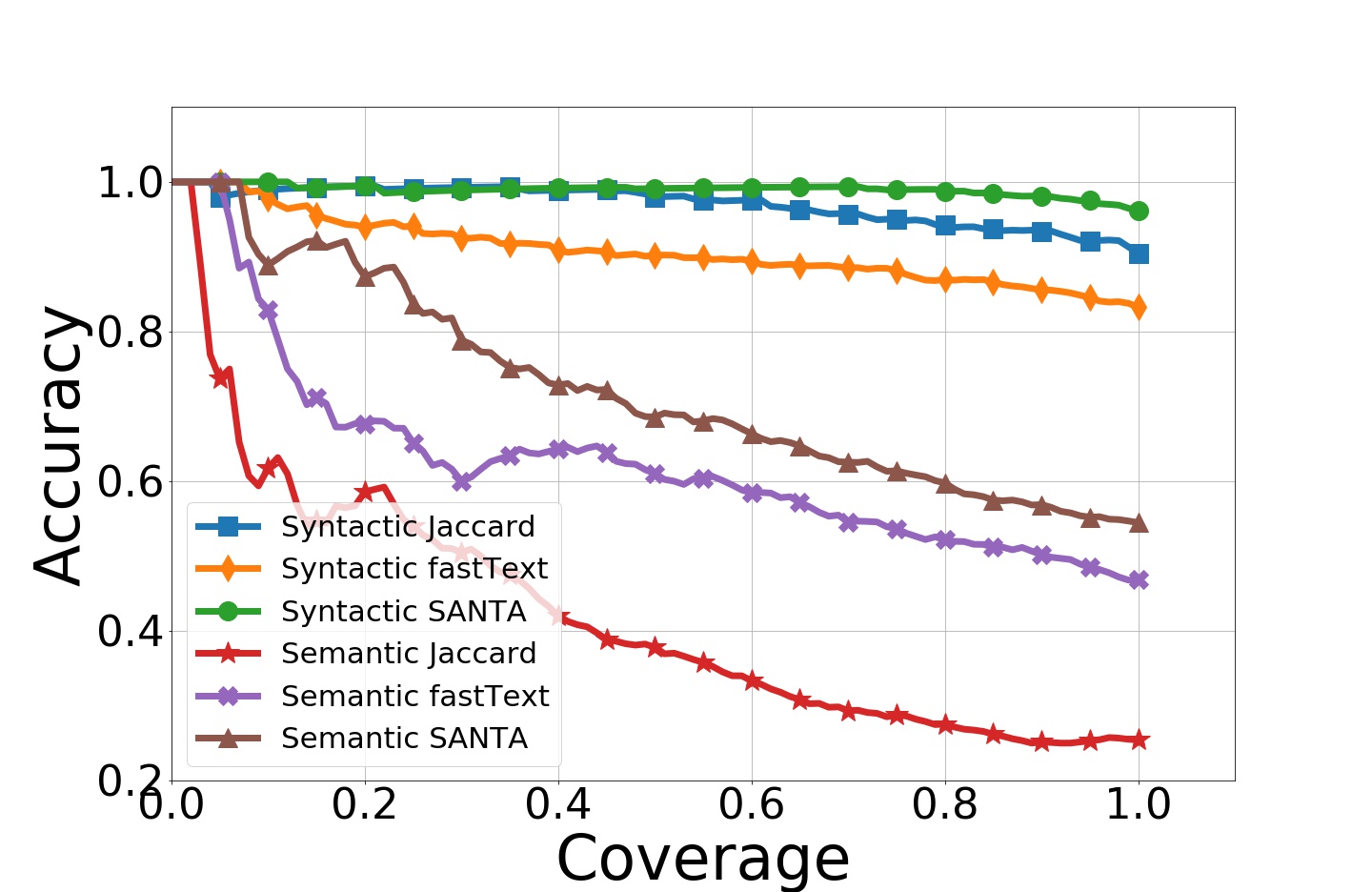}}
		\caption{Study of various normalization algorithms on semantic and syntactic dataset}
		\label{fig:semanticsyntactic}
	\end{center}
	\vspace{-7mm}
\end{figure}

\vspace{-2mm}
\subsection{Study on Syntactic and Semantic Dataset}
\label{semanticsyntactic}
\vspace{-2mm}

In this section, we do a separate comparison of normalization algorithms on samples requiring semantic and syntactic matching. 
We filtered test dataset where true label is not `Others', and manually labelled each surface form as requiring syntactic or 
semantic similarity. Based on this analysis, we observe that $45\%$ of test data requires syntactic matching, $17\%$ requires 
semantic matching and remaining $38\%$ is mapped to `other' class. For current analysis of syntactic and semantic set, we use
a special case of metric defined in section \ref{sec:exp} (since `other' class is not present). We set $x_{1} = 0$, ensuring
that `other' class is not predicted for any samples of test data. We show Accuracy-Coverage plot for semantic and syntactic cases in 
Figure \ref{fig:semanticsyntactic}.

For semantic set, we observe that fastText performs better than string similarity, due to its ability to learn semantic representation. 
Our proposed SANTA framework, further improves over fastText for better semantic matching with close canonical forms. For syntactic set, we observe comparable 
performance of SANTA and string similarity. These results demonstrate that our proposed SANTA framework performs well on both syntactic 
and semantic set.

\vspace{-3mm}
\subsection{Word Embeddings Visualization}
\label{embedd_visual}
\begin{figure*}[htbp]
	\begin{subfigure}{\columnwidth}
		\caption{\textbf{FastText: Headphone Color}}
		\includegraphics[width=\linewidth, height=7cm]{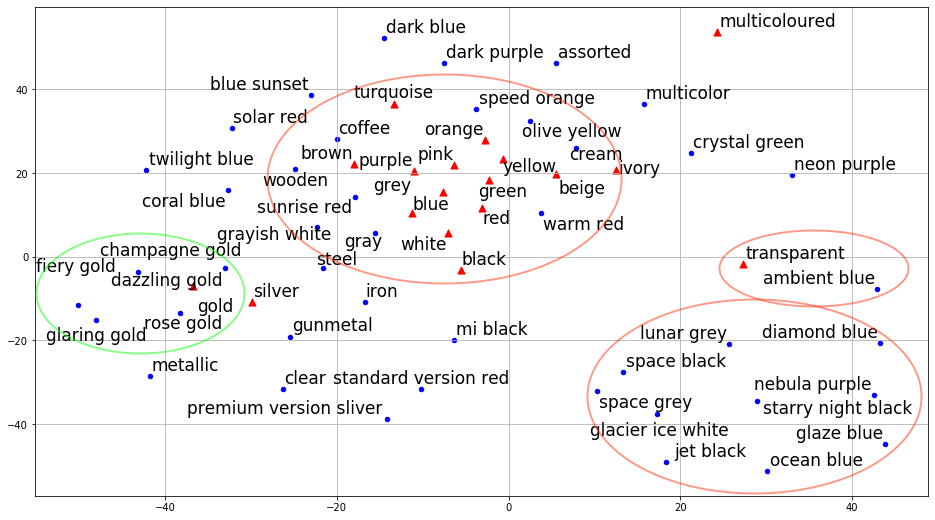}
		\label{FT_only_color}
	\end{subfigure}
	\begin{subfigure}{\columnwidth}
		\caption{\textbf{SANTA: Headphone Color}}
		\includegraphics[width=\linewidth, height=7cm]{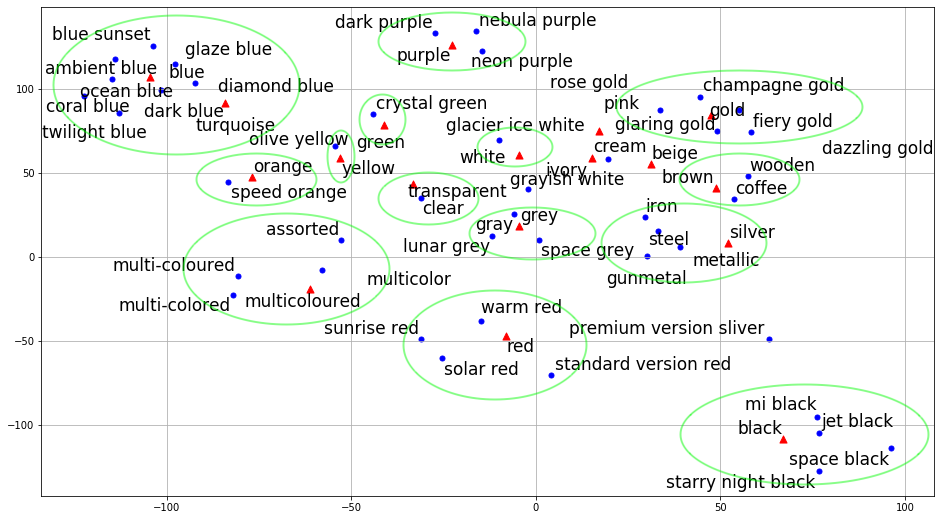}
		\label{TP_only_color}
	\end{subfigure}
	\begin{subfigure}{\columnwidth}
		\caption{\textbf{FastText: Necklace type}}
		\includegraphics[width=\linewidth, height=6cm]{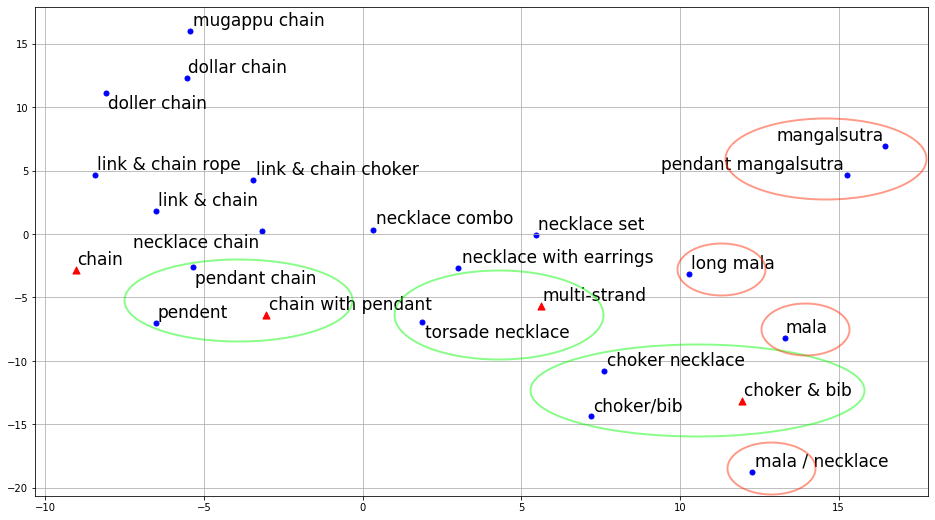}
		\label{FT_only_necklacetype}
	\end{subfigure}
	\begin{subfigure}{\columnwidth}
		\caption{\textbf{SANTA: Necklace type}}
		\includegraphics[width=\linewidth, height=6cm]{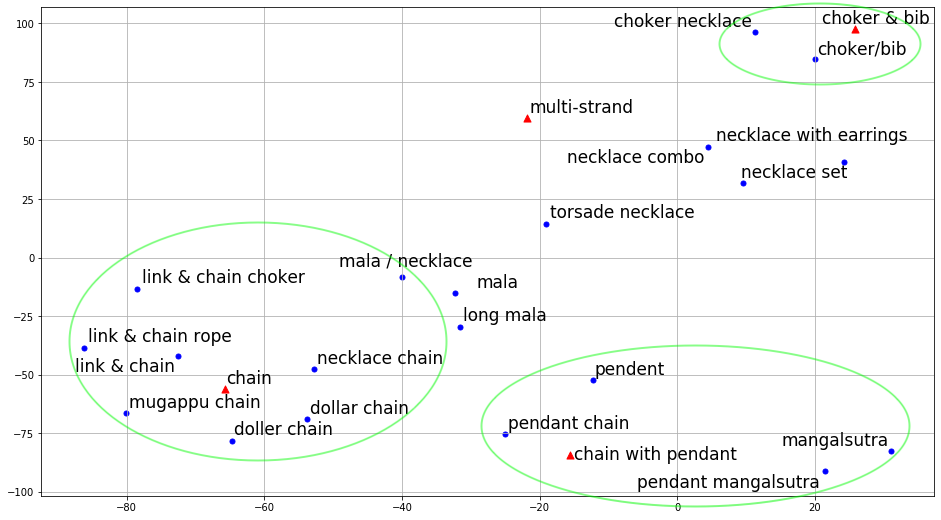}
		\label{TP_only_necklacetype}
	\end{subfigure}
	\begin{subfigure}{\columnwidth}
		\caption{\textbf{FastText: Watch movement type}}
		\includegraphics[width=\linewidth, height=6cm]{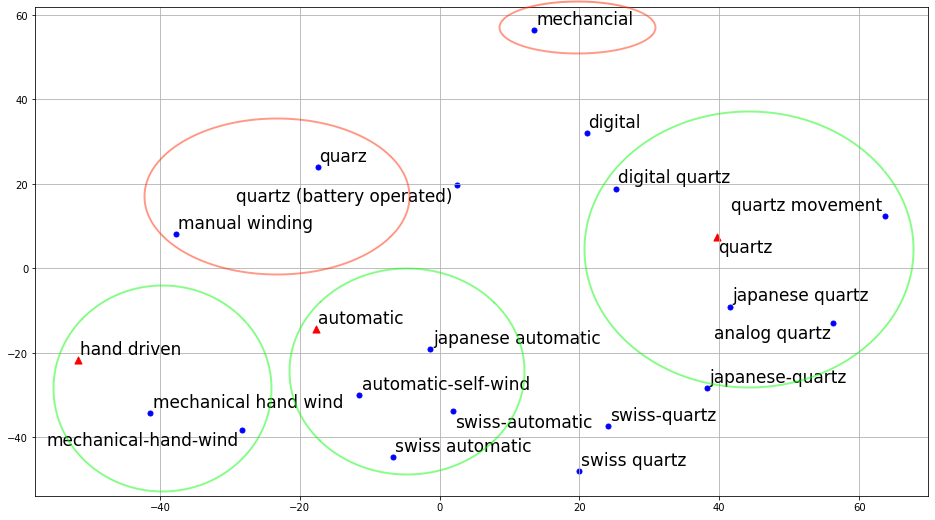}
		\label{FT_only_watchmovementtype}
	\end{subfigure}
	\begin{subfigure}{\columnwidth}
		\caption{\textbf{SANTA: Watch movement type}}
		\includegraphics[width=\linewidth, height=6cm]{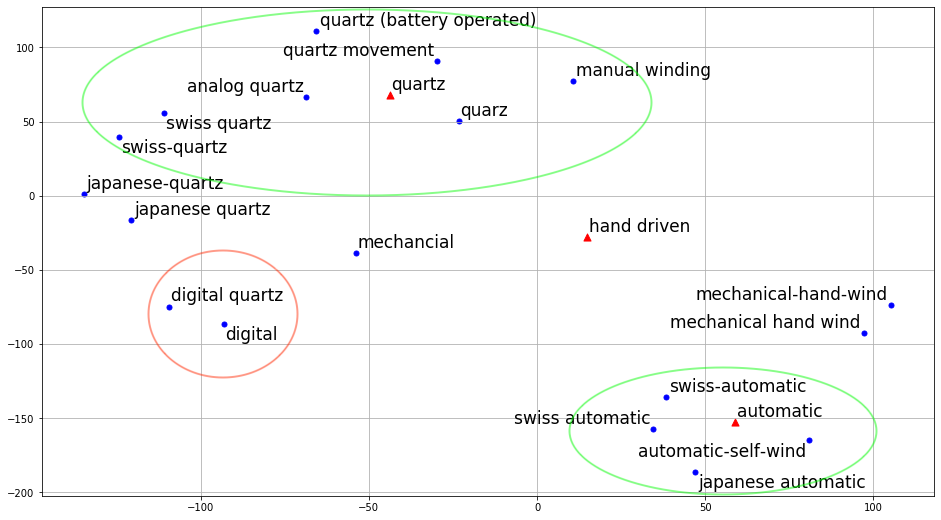}
		\label{SANTA}
	\end{subfigure}
	\caption{Figure showing t-SNE plot of fastText and SANTA embeddings for three attributes. Surface forms are shown with green dots and canonical forms with red triangles. 
	For better understanding of results, we use green oval selection to show correct homogenous cohorts and red oval selection for incorrect cohorts. This figure is best seen in colors.}
	\label{fig:tsne_plot}
\end{figure*}

\vspace{-2mm}
For qualitative comparison of fastText and SANTA embeddings, we project these embeddings into 2-dimensions using 
t-SNE~\cite{JMLR:v9:vandermaaten08a}. Figure~\ref{fig:tsne_plot} shows t-SNE plots\footnote{\url{https://scikit-learn.org/stable/modules/generated/sklearn.manifold.TSNE}} for 3 attributes (Headphone Color, Jewelry Necklace type 
and Watch Movement type). For color attribute, we observe that values based on SANTA have homogenous cohorts of canonical values and corresponding surface forms
(e.g. there is a cohort for `black' color on bottom-right and `blue' color on top-left of the plot.). However, with fastText, the color values are scattered across the plot 
without any specific cohorts. Similar patterns are seen with necklace type where SANTA results show better cohorts than fastText. These results demonstrate that embeddings 
learnt with SANTA are better suited than fastText embeddings to distinguish between close canonical forms.

\section{Conclusion}
\vspace{-2mm}
\label{sec:conclusions}
In this paper, we studied the problem of attribute normalization for E-commerce. We did a systematic study of multiple syntactic matching 
algorithms and established that use of `cosine similarity' leads to $2.7\%$ improvement over commonly used Jaccard index. Additionally, we argued 
that attribute normalization requires combination of syntactic and semantic matching. We described our SANTA framework for attribute normalization, including our
proposed task to learn embeddings in a self-supervised fashion with twin network and triplet loss. Evaluation on a real-world dataset for $50$ attributes, shows
that embeddings learnt using our proposed SANTA framework outperforms best string matching algorithm by $2.3\%$ and fastText by $19.3\%$ for attribute normalization task. 
Our evaluation based on semantic and syntactic examples and t-SNE plots provide useful insights into qualitative behaviour of these embeddings.

\bibliographystyle{acl_natbib}
\bibliography{anthology,rf}

\newpage

\section{Appendix}
\label{sec:appendix}

We list few interesting examples in Table~\ref{tab:qualitative}. It can be observed that string similarity makes correct predictions for cases requiring fuzzy matching
(e.g. matching ``\textit{multi}" with ``\textit{multicoloured}"), but, makes incorrect predictions for examples requiring semantic matching (e.g. 
incorrectly matching ``\textit{cane}" with ``\textit{polyurethane}"). With fastText, we get correct predictions for many semantic examples, however, 
we get incorrect predictions for close canonical forms (e.g. incorrectly mapping ``\textit{3 seater sofa set}" to ``\textit{five seat}" as 
``\textit{three seat}" and ``\textit{five seat}" occur in similar contexts in title). Our proposed SANTA model does well for most of these examples,
but it fails to make correct predictions for rare surface forms (e.g. ``\textit{no assembly required, pre-aseembled}").

\begin{table}[!htbp]
	\begin{adjustwidth}{-14mm}{}
		\begin{tabular}{|c|c|c|c|c|c|c|}
			\hhline{-|-|-|-|-|-|} \hline
			Surface Form                        & \makecell{ Actual \\Canonical Form} & \makecell{Cosine Similarity\\ Prediction}              & \makecell{FastText \\ Prediction}                                    & \makecell{SANTA \\prediction}                                                & Comment                                                                  \\\hhline{-|-|-|-|-|-|}
			\hline
			multi                               & multicoloured       & \cellcolor[HTML]{9AFF99}multicoloured     & \cellcolor[HTML]{FFCCC9}green                          & \cellcolor[HTML]{9AFF99}{\color[HTML]{000000} multicoloured}    &    \\\hhline{-|-|-|-|-|}
			thermoplastic                               & plastic       & \cellcolor[HTML]{9AFF99}plastic     & \cellcolor[HTML]{FFCCC9}silicone                          & \cellcolor[HTML]{9AFF99}{\color[HTML]{000000} plastic}    &   \\\hhline{-|-|-|-|-|}
			amd radeon r3                               & ati radeon       & \cellcolor[HTML]{9AFF99}ati radeon     & \cellcolor[HTML]{FFCCC9}nvidia geforce                          & \cellcolor[HTML]{9AFF99}{\color[HTML]{000000} ati radeon}    &   \\\hhline{-|-|-|-|-|}
			free size                               & one size       & \cellcolor[HTML]{9AFF99}one size     & \cellcolor[HTML]{FFCCC9}small                          & \cellcolor[HTML]{9AFF99}{\color[HTML]{000000} one size}    & \multirow{-4}{*} {\makecell{FastText fails}}    \\ \hhline{-|-|-|-|-|-|}
			\hline
			2 years                             & 2 - 3 years         & \cellcolor[HTML]{FFCCC9}11 - 12 years     & \cellcolor[HTML]{FFCCC9}3 - 4 years                    & \cellcolor[HTML]{9AFF99}{\color[HTML]{000000} 2 - 3 years}      &    \multirow{7}{*}{\makecell{Both String \\Similarity and \\fastText fails \\but SANTA \\gives correct\\ mapping}}                                                                                                \\ \hhline{-|-|-|-|-|~|}
			elbow sleeve                        & half sleeve         & \cellcolor[HTML]{FFCCC9}3/4 sleeve        & \cellcolor[HTML]{FFCCC9}short sleeve                   & \cellcolor[HTML]{9AFF99}{\color[HTML]{000000} half sleeve}      &                                                                                                         \\ \hhline{-|-|-|-|-|~|}
			cane                                & bamboo              & \cellcolor[HTML]{FFCCC9}polyurethane      & \cellcolor[HTML]{FFCCC9}rattan                         & \cellcolor[HTML]{9AFF99}{\color[HTML]{000000} bamboo}           &                                                                                                         \\ \hhline{-|-|-|-|-|~|}
			\makecell{product will \\be assembled}           & \makecell{requires\\ assembly}   & \cellcolor[HTML]{FFCCC9}\makecell{already assembled} & \cellcolor[HTML]{FFCCC9}d-i-y                          & \cellcolor[HTML]{9AFF99}{\color[HTML]{000000} \Gape{\makecell{require\\ assembly}}} &                                                                                                         \\
			\hhline{-|-|-|-|-|~|}
			\makecell{3 seater \\sofa set}                   & three seat          & \cellcolor[HTML]{FFCCC9}four seat         & \cellcolor[HTML]{FFCCC9}five seat                      & \cellcolor[HTML]{9AFF99}{\color[HTML]{000000} three seat}       &  \\ \hhline{-|-|-|-|-|-|}
			\hline
			nokia os                            & symbian             & \cellcolor[HTML]{FFCCC9}palm web os       & \cellcolor[HTML]{9AFF99}{\color[HTML]{000000} symbian} & \cellcolor[HTML]{9AFF99}{\color[HTML]{000000} symbian}          &                                                                                                         \\
			\hhline{-|-|-|-|-|~|}
			silicone                            & rubber              & \cellcolor[HTML]{FFCCC9}silk              & \cellcolor[HTML]{9AFF99}{\color[HTML]{000000} rubber}  & \cellcolor[HTML]{9AFF99}{\color[HTML]{000000} rubber}           &                                                                                                         \\
			\hhline{-|-|-|-|-|~|}
			coffee                              & brown               & \cellcolor[HTML]{FFCCC9}off-white         & \cellcolor[HTML]{9AFF99}{\color[HTML]{000000} brown}   & \cellcolor[HTML]{9AFF99}{\color[HTML]{000000} brown}            & \multirow{-3}{*}{\makecell{String Similarity \\fails}}    \\ \hhline{-|-|-|-|-|-|}
			\hline
			\makecell{no assembly \\required, \\pre-aseembled} & \makecell{already\\ assembled}   & \cellcolor[HTML]{FFCCC9}requires assembly & \cellcolor[HTML]{9AFF99}already assembled              & \cellcolor[HTML]{FFCCC9} \Gape{\makecell{requires\\ assembly}}                   &                                                                                    \\ \hhline{-|-|-|-|-|~|}
			mechancial                          & hand driven         & \cellcolor[HTML]{9AFF99}hand driven       & \cellcolor[HTML]{9AFF99}hand driven                    & \cellcolor[HTML]{FFCCC9}automatic                               &\multirow{-4}{*}{\makecell{SANTA fails}}\\ \hhline{-|-|-|-|-|-|}
			\hline                                          
		\end{tabular}
	\end{adjustwidth}
	\caption{Qualitative Examples for multiple normalization approaches. Correct predictions are highlighted in green color and incorrect predictions are highlighted in red color.}
	\label{tab:qualitative}
\end{table}

\end{document}